\documentclass[final]{cvpr}

\usepackage[table]{xcolor}
\usepackage[utf8]{inputenc}
\usepackage{times}
\usepackage{epsfig}
\usepackage{graphicx}
\usepackage{amsmath}
\usepackage{amssymb}
\usepackage{multirow}
\usepackage{booktabs}
\usepackage{hhline}
\usepackage{subcaption}
\usepackage{eucal}

\definecolor{rowblue}{RGB}{220,230,240}
\usepackage[pagebackref=true,breaklinks=true,colorlinks,bookmarks=false]{hyperref}

\def\cvprPaperID{4667} % *** Enter the CVPR Paper ID here
\def\confYear{CVPR 2021}

\newcommand{\comment}[1]{}

\newcommand{\bzero}{\textbf{0}}

\newcommand{\bc}{\mathbf{c}}
\newcommand{\bd}{\mathbf{d}}

\newcommand{\bo}{\mathbf{o}}
\newcommand{\bp}{\mathbf{p}}

\newcommand{\bx}{\mathbf{x}}

\newcommand{\bI}{\mathbf{I}}

\newcommand{\bT}{\mathbf{T}}

\newcommand{\mM}{\mathcal{M}}

\newcommand{\mL}{\mathcal{L}}

\newcommand{\mT}{\mathcal{T}}

\newcommand{\bDelta}{\boldsymbol{\Delta}}

\let\oldtwocolumn\twocolumn
\renewcommand\twocolumn[1][]{%
    \oldtwocolumn[{#1}{
    \begin{center}
          \vspace{-3em}
           \includegraphics[width=\textwidth]{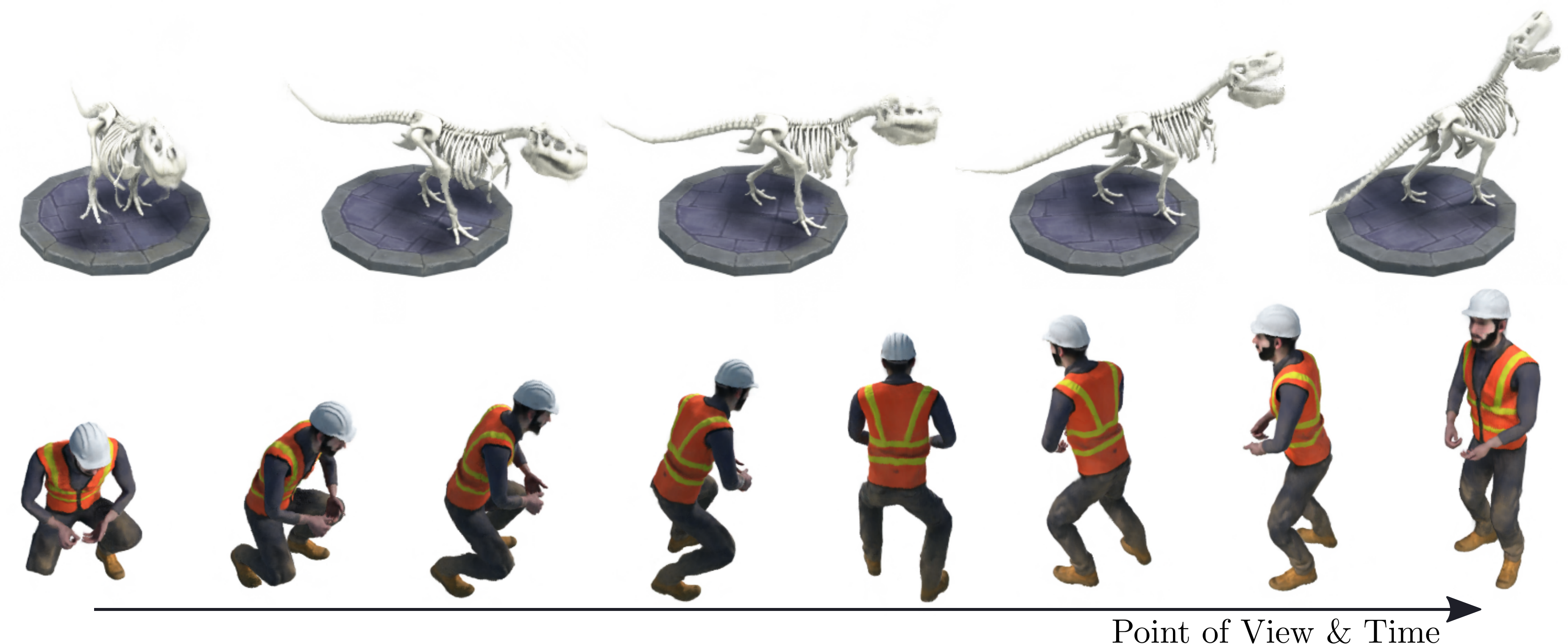}
    \vspace*{-1.75em} 
    \captionof{figure}{\small{We propose D-NeRF, a method for synthesizing novel views, at an arbitrary point in time, of dynamic scenes with complex non-rigid geometries. We optimize an underlying deformable volumetric function from a sparse set of input monocular views without the need of ground-truth geometry nor multi-view images. The figure shows two scenes under variable points of view and time instances synthesised by the proposed model.} }
    \label{fig:teaser}
        \end{center}
    }]
}

\begin{document}

%%%%%%%%% TITLE
\title{D-NeRF: Neural Radiance Fields for Dynamic Scenes}

\author{Albert Pumarola$^{1}$\hspace{0.8cm}
Enric Corona$^{1}$ \hspace{0.8cm}
Gerard  Pons-Moll$^{2}$ \hspace{0.8cm}
Francesc  Moreno-Noguer$^{1}$ \hspace{0.8cm}\\
$^{1}$Institut de Rob\`otica i Inform\`atica Industrial, CSIC-UPC\\
$^{2}$Max Planck Institute for Informatics
}

\maketitle
%\thispagestyle{empty}

%%%%%%%%% ABSTRACT
\begin{abstract}
Neural rendering techniques combining machine learning  with geometric reasoning have arisen as one of the most promising  approaches for  synthesizing novel views of a scene from a sparse set of images. 
Among these, stands out the Neural radiance fields (NeRF)~\cite{mildenhall2020nerf}, which trains a deep  network to map  5D input coordinates  (representing spatial location and viewing direction) into a volume density and view-dependent emitted radiance. However, despite achieving an unprecedented level of photorealism on the generated images, NeRF  is only applicable to static scenes, where the same spatial location can be queried from different images. 
In this paper we introduce D-NeRF, a method that extends neural radiance fields to a dynamic domain, allowing to reconstruct and render novel images of objects under rigid and non-rigid motions from a \emph{single} camera moving around the scene. 
For this purpose we consider time as an additional  input to the system, and split the learning process in two main stages: one that encodes the scene into a canonical space and another that maps this canonical representation into the deformed scene at a particular time. Both mappings are  simultaneously learned using fully-connected networks. Once the networks are trained, D-NeRF can render novel images, controlling both the camera view and the time variable, and thus, the object movement. We demonstrate the effectiveness of our approach on scenes with objects under rigid, articulated and non-rigid motions. Code, model weights and the dynamic scenes dataset will be released.

\end{abstract}

%%%%%%%%% BODY TEXT

\section{Introduction}
Rendering novel photo-realistic views of a scene from a sparse set of input images is necessary for many applications in \eg augmented reality, virtual reality, 3D content production, games and the movie industry.
Recent advances in the emerging field of neural rendering, which learn scene representations encoding both geometry and appearance~\cite{mildenhall2020nerf,martin2020nerf,liu2020neural,yariv2020multiview,niemeyer2020differentiable,rematas2020neural}, have achieved results that largely surpass those  of traditional Structure-from-Motion~\cite{hartley2003multiple,triggs1999bundle,snavely2006photo}, light-field photography~\cite{levoy1996light} and image-based rendering approaches~\cite{buehler2001unstructured}. For instance, the Neural Radiance Fields (NeRF)~\cite{mildenhall2020nerf} have shown that simple multilayer perceptron networks can encode the mapping from 5D inputs (representing spatial locations $(x,y,z)$ and camera views $(\theta,\phi)$) to  emitted radiance values and volume density. This learned mapping allows then free-viewpoint rendering with extraordinary realism. Subsequent works have extended Neural Radiance Fields to images in the wild undergoing severe lighting changes~\cite{martin2020nerf} and have proposed sparse voxel fields for rapid inference~\cite{liu2020neural}. Similar schemes have also been recently used for multi-view surface reconstruction~\cite{yariv2020multiview} and learning surface light fields~\cite{oechsle2020learning}.

Nevertheless, all these approaches assume a \emph{static} scene without moving objects. 
In this paper we relax this assumption and propose, to the best of our knowledge, the first end-to-end neural rendering system that is applicable to dynamic scenes, made of both still and moving/deforming objects. While there exist approaches for 4D view synthesis~\cite{BansalCVPR2020}, our approach is different in that: 1) we only require a single camera; 2) we do not need to pre-compute a 3D reconstruction; and 3) our approach can be trained end-to-end.

Our idea is to represent the input of our system with a continuous 6D function, which besides 3D location and camera view, it also considers the time component $t$. Naively extending NeRF to learn a mapping from $(x,y,z,t)$ to density and radiance does not produce satisfying results, as the temporal redundancy in the scene is not effectively exploited. 
Our observation is that objects can move and deform, but typically do not appear or disappear.
Inspired by classical 3D scene flow~\cite{vedula2005three}, the core idea to build our method, denoted Dynamic-NeRF (D-NeRF in short), is to decompose learning in two modules. The first one learns a spatial mapping $(x,y,z,t)\rightarrow(\Delta x, \Delta y, \Delta z)$ between each point of the scene at time $t$ and a \emph{canonical scene} configuration. The second module regresses the scene radiance emitted in each direction and volume density given the tuple $(x+\Delta x,y+\Delta y,z+\Delta z,\theta, \phi)$. Both mappings are learned with deep fully connected networks without convolutional layers. The learned model then allows to synthesize novel images, providing control in the continuum $(\theta, \phi, t)$ of the camera views and time component, or equivalently, the dynamic state of the scene (see Fig.~\ref{fig:teaser}).

We thoroughly evaluate D-NeRF on scenes undergoing very different types of deformation, from articulated motion to humans performing complex body poses. We show that by decomposing learning into a canonical scene and scene flow  D-NeRF is able to render high-quality images while controlling both camera view and time components. As a side-product, our method is also able to produce complete 3D  meshes that capture the time-varying geometry  and which remarkably are obtained by  observing the scene under a specific deformation only from one single viewpoint.

\begin{figure*}
    \centering
    \includegraphics[width=\linewidth]{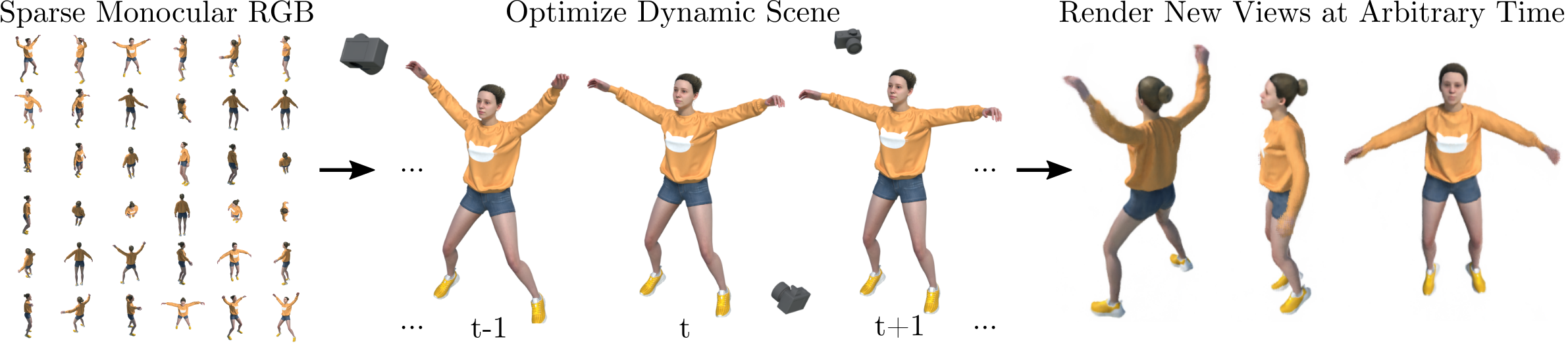}
    \vspace*{-1.75em} 
    \caption{\small{\textbf{Problem Definition.} Given a sparse set of images of a dynamic scene moving non-rigidly and being captured by a monocular camera,  we aim to design a deep learning model  to implicitly encode the scene and synthesize novel views at an arbitrary time. Here, we visualize a subset of the input training frames paired with accompanying camera parameters, and we show three novel views at three different time instances rendered by the proposed method.}}
    \vspace{-1.75em}
    \label{fig:pipeline}
\end{figure*}

\section{Related work}

\paragraph{Neural implicit representation for 3D geometry.}
The success of deep learning on the 2D domain has spurred a growing interest in the 3D domain. Nevertheless, which is the most appropriate 3D data representation for deep learning remains an open question, especially for non-rigid geometry. Standard representations for rigid geometry include point-clouds~\cite{SuCVPR2017,pumarola2020c}, voxels~\cite{GirdharECCV2016,YanNIPS2016} and octrees~\cite{WangSIGGRAPH2017,TatarchenkoICCV2017}. Recently, there has been a strong burst in representing 3D data in an implicit manner via a neural network~\cite{mescheder2019occupancy,park2019deepsdf,chen2019learning,xu2019disn,chibane2020implicit,genova2020local}. The main idea behind this approach is to describe the information (\eg occupancy, distance to surface, color, illumination) of a 3D point $\bx$ as the output of a neural network $f(\bx)$. Compared to the previously mentioned representations, neural implicit representations allow for continuous surface reconstruction at a low memory footprint. 

The first works exploiting implicit representations~\cite{mescheder2019occupancy, park2019deepsdf,chen2019learning,xu2019disn} for  3D representation were limited by their requirement of having access to 3D ground-truth geometry, often expensive or even impossible to obtain for in the wild scenes. Subsequent works relaxed this requirement by introducing a differentiable render allowing 2D supervision. For instance, \cite{liu2019learning} proposed an efficient ray-based field probing algorithm for efficient image-to-field supervision. \cite{niemeyer2020differentiable,yariv2020universal} introduced an implicit-based method to calculate the exact derivative of a 3D occupancy field surface intersection with a camera ray. In~\cite{sitzmann2019scene}, a recurrent neural network was used to ray-cast the scene and estimate the surface geometry. However, despite these techniques have a great potential to represent 3D shapes in an unsupervised manner, they are typically limited to relatively simple geometries. 

NeRF~\cite{mildenhall2020nerf} showed that by implicitly representing a rigid scene using 5D radiance fields makes it possible to capture high-resolution geometry and photo-realistically rendering  novel views. \cite{martin2020nerf} extended this method to handle variable illumination and transient occlusions to deal with in the wild images. In~\cite{liu2020neural}, even more complex 3D surfaces were represented by using voxel-bouded implicit fields. And~\cite{yariv2020multiview} circumvented the need of multiview camera calibration.

However, while all mentioned methods achieve impressive  results on rigid scenes, none of them can deal with dynamic and deformable scenes. Occupancy flow~\cite{niemeyer2019occupancy} was the first work to tackle non-rigid geometry by learning continuous vector field assigning a motion vector to every point in space and time, but it requires full 3D  ground-truth supervision.  Neural volumes~\cite{lombardi2019neural} produced high quality reconstruction results via an encoder-decoder voxel-based representation enhanced with an implicit voxel warp field, but they require a muti-view image capture setting.

To the best of our knowledge, D-NeRF  is the first approach able to generate a neural implicit representation for non-rigid and time-varying scenes, trained solely on monocular data without the need of 3D ground-truth supervision nor a multi-view camera setting.

\paragraph{Novel view synthesis.}
Novel view synthesis is a long standing vision and graphics problem that aims to synthesize new images from arbitrary view points of a scene captured by multiple images. Most traditional approaches for rigid scenes consist on reconstructing the scene from multiple views with Structure-from-Motion~\cite{hartley2003multiple} and bundle adjustment~\cite{triggs1999bundle}, while other approaches propose light-field based photography~\cite{levoy1996light}. More recently, deep learning based techniques~\cite{shen2019patient,kar2017learning,flynn2019deepview,choi2019extreme,mildenhall2019llff} are able to learn a neural volumetric representation from a set of sparse images.

However, none of these methods can synthesize novel views of dynamic scenes. To tackle non-rigid scenes most methods approach the problem by reconstructing a dynamic 3D textured mesh. 3D reconstruction of non-rigid surfaces from monocular images is known to be severely ill-posed. Structure-from-Template (SfT) approaches~\cite{bartoli2015shape,chhatkuli2014stable,moreno2013pami} recover the surface geometry given a reference known template configuration. Temporal information is another prior typically exploited. Non-rigid-Structure-from-Motion (NRSfM) techniques~\cite{tomasi1992shape,agudo2015simultaneous} exploit temporal information. Yet,   SfT and NRSfM require either  2D-to-3D matches or 2D point tracks,  limiting their general applicability to relatively well-textured surfaces and mild deformations.

Some of these limitations are overcome by learning based techniques, which have  been effectively used for synthesizing  novel photo-realistic views of dynamic scenes. For instance, \cite{BansalCVPR2020,zitnick2004high,jiang20123d}  capture the dynamic scene at the same time instant from multiple views, to then generate 4D space-time visualizations. \cite{flynn2016deepstereo,philip2018plane,zhou2018stereo} also leverage on simultaneously capturing the scene from  multiple cameras to estimate  depth,  completing areas with missing information and then performing view synthesis.   In~\cite{yoon2020novel}, the need of multiple views is circumvented by using a pre-trained network that estimates a per frame depth. This depth, jointly with the optical flow and consistent depth estimation across frames, are then used to interpolate between images and render novel views. Nevertheless, by decoupling depth estimation from novel view synthesis, the outcome of this approach becomes highly dependent on the quality of the depth maps as well as on the reliability of the optical flow. Very recently, X-Fields~\cite{bemana2020x} introduced a neural network to interpolate between images taken across different view, time or illumination conditions. However, while this approach is able to process dynamic scenes, it requires more than one view. Since no 3D representation is learned, variation in viewpoint is small.

D-NeRF is different from all prior work in that it does not require 3D reconstruction, can be learned end-to-end, and requires \emph{a single view} per time instance.
Another appealing characteristic of D-NeRF is that  it inherently learns a time-varying 3D volume density and emitted radiance, which turns the novel view synthesis into a ray-casting process instead of a view interpolation, which is remarkably more robust to rendering images from arbitrary viewpoints.

\begin{figure*}
    \centering
    \includegraphics[width=\linewidth]{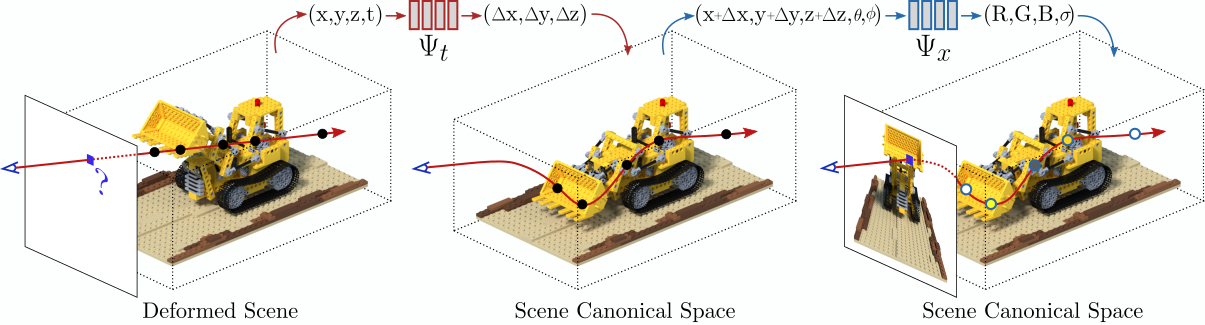}
    \vspace*{-2em} 
    \caption{\small{\textbf{D-NeRF Model}. The proposed architecture consists of two main blocks: a deformation network $\Psi_t$ mapping all scene deformations to a common canonical configuration; and a canonical network $\Psi_x$ regressing volume density and view-dependent RGB color from every camera ray.
    }}
    \vspace{-1.75em} 
    \label{fig:Model}
\end{figure*}

\section{Problem Formulation}
Given a sparse set of images of a dynamic scene captured with a monocular camera,  we aim to design a deep learning model able to implicitly encode the scene and synthesize novel views at an arbitrary time (see Fig.~\ref{fig:pipeline}).

Formally, our goal is to learn a mapping $\mM$ that, given  a 3D point $\bx=(x, y, z)$, outputs its emitted color $\bc=(r,g,b)$ and volume density $\sigma$ conditioned on a time instant $t$ and view direction $\bd = (\theta, \phi)$. That is, we seek to estimate the mapping $\mM: (\bx, \bd, t) \rightarrow (\bc, \sigma)$.

An intuitive  solution would be to directly learn the transformation $\mM$ from the 6D space $(\bx, \bd, t)$ to the 4D space $(\bc, \sigma)$. However, as we will show in the results section, we obtain consistently better results by splitting the mapping  $\mM$
into $\Psi_x$ and $\Psi_t$, where $\Psi_x$ represents the scene in canonical configuration and $\Psi_t$ a mapping between the scene at time instant $t$ and the canonical one. More precisely, given a point $\bx$ and viewing direction $\bd$ at time instant $t$ we first transform the point position to its canonical configuration as $\Psi_t: (\bx, t) \rightarrow \Delta\bx$. Without loss of generality, we chose $t=0$ as the canonical scene $\Psi_t: (\bx, 0) \rightarrow \bzero$. 
By doing so the scene is no longer independent between time instances, and becomes interconnected through a common canonical space anchor.
Then, the assigned emitted color and volume density  under viewing direction $\bd$ equal to those in the canonical configuration  $\Psi_x: (\bx+\Delta\bx,\bd) \rightarrow (\bc, \sigma)$. 
 
We propose to learn $\Psi_x$ and $\Psi_t$  using  a sparse set of $T$  RGB images $\{ \bI_t, \bT_t\}_{t=1}^T$ captured with a monocular camera, where $\bI_t \in \mathbb{R}^{H \times W \times 3}$ denotes the image acquired under camera   pose    $\mathbf{\bT}_t \in \mathbb{R}^{4 \times 4}$ SE(3), at time $t$. Although we could assume multiple views per time instance, we want to test the limits of our method, and assume a single image per time instance. That is, we do not observe the scene under a specific configuration/deformation state from different viewpoints. 
\section{Method}
We now introduce D-NeRF, our novel neural renderer for view synthesis  trained solely from a sparse set of images of a dynamic scene. We build on NeRF~\cite{mildenhall2020nerf} and generalize it to handle non-rigid scenes. Recall that NeRF requires multiple views of a rigid scene
In contrast, D-NeRF can learn a volumetric density representation for continuous non-rigid scenes trained with a single view per time instant.

As shown in Fig.~\ref{fig:Model}, D-NeRF consists of two main neural network modules, which parameterize the mappings explained in the previous section $\Psi_t,\Psi_x$.
On the one hand we have the {\em Canonical Network}, an MLP (multilayer perceptron) $\Psi_x(\bx,\bd)\mapsto (\bc, \sigma)$ is trained to encode the scene in the canonical configuration such that given a 3D point $\bx$ and a view direction   $\bd$ returns its emitted color $\bc$ and volume density $\sigma$. The second module is called {\em Deformation Network} and consists of another MLP $\Psi_t(\bx,t)\mapsto \Delta\bx$ which predicts a deformation field defining the transformation between the scene at time $t$ and the scene in its canonical configuration. We next describe in detail each one of these blocks (Sec.~\ref{sec:architecture}), their interconnection for volume rendering (Sec.~\ref{sec:render}) and how are they learned (Sec.~\ref{sec:learn}).

\subsection{Model Architecture}
\label{sec:architecture}

\vspace{1mm} 
\noindent{\bf Canonical Network.}  With the use of a canonical configuration we seek to find a representation of the scene that brings together the information of all corresponding points in all images. By doing this, the missing information from a specific viewpoint can then be retrieved from that canonical configuration, which shall act as an anchor interconnecting all images. 

The canonical  network $\Psi_x$ is  trained so as to encode volumetric density and color of the scene in canonical configuration. Concretely, given the 3D coordinates $\bx$ of a point, we first encode it into a    256-dimensional feature vector. This feature vector is then concatenated with the camera viewing direction  $\bd$, and propagated through a fully connected layer to yield   the emitted color $\bc$ and volume density $\sigma$ for that given point in the canonical space.

\vspace{1mm} 
\noindent{\bf Deformation Network.}
The deformation network $\Psi_t$ is optimized to estimate the deformation field between the scene at a specific time instant and the scene in canonical space. Formally, given a 3D point $\bx$ at time $t$, $\Psi_t$ is trained to output the displacement $\Delta\bx$ that transforms the given point to its position in the canonical space as $\bx+\Delta\bx$. For all experiments, without loss of generality, we set the canonical scene to be the scene at time $t=0$:
\begin{equation}
\Psi_t(\bx, t) = \begin{cases} \Delta\bx, & \mbox{if } t \neq 0 \\
0, & \mbox{if } t = 0 \end{cases}
\label{eq:deformation}
\end{equation}

As shown in previous works~\cite{rahaman2019spectral,vaswani2017attention,mildenhall2020nerf}, directly feeding raw coordinates and angles to a neural network results in low performance. Thus, for both the canonical and the deformation networks, we first encode $\bx$, $\bd$ and $t$ into a higher dimension space. We use the same positional encoder as in~\cite{mildenhall2020nerf} where $\gamma(p) = <(\sin(2^l\pi p), \cos(2^l\pi p))>_0^L$. We independently apply the encoder $\gamma(\cdot)$ to each coordinate and camera view component, using  $L=10$ for $\bx$, and $L=4$ for $\bd$ and $t$.

\subsection{Volume Rendering}
\label{sec:render}
We now adapt NeRF volume rendering equations to account for non-rigid deformations in the proposed 6D neural radiance field. Let $\bx(h)=\bo+h\bd$ be a point along the camera ray emitted from the center of projection $\bo$ to a pixel $p$. Considering near and far bounds $h_n$ and $h_f$ in that ray, the expected color $C$ of the pixel $p$ at time $t$ is given by: 
\begin{align}
    C(p,t) = \int_{h_n}^{h_f}\mT(&h,t)\sigma(\bp(h,t))\bc(\bp(h,t), \bd)dh, \label{eq:integral} \\
    \quad \text{where}
     \quad \bp(h,t) &= \bx(h)+\Psi_t(\bx(h), t),\\
    \quad [\bc(\bp(h,t), \bd), \sigma&(\bp(h,t))] = \Psi_x(\bp(h,t), \bd), \\
    \quad \text{and} 
    \quad \mT(h, t) = &\exp\left(-\int_{h_n}^{h} \sigma(\bp(s,t)) ds \right).
    \label{eq:transmitance}
\end{align}
The 3D point $\bp(h,t)$ denotes the  point  on the camera ray $\bx(h)$ transformed to canonical space using our Deformation Network $\Psi_t$, and $\mT(h, t)$ is the accumulated probability that the ray emitted from $h_n$ to $h_f$ does not hit any other particle. Notice that the density $\sigma$ and color $\bc$ are predicted by our Canonical Network $\Psi_x$.

As in~\cite{mildenhall2020nerf} the volume rendering integrals in Eq.~\eqref{eq:integral} and Eq.~\eqref{eq:transmitance} can be approximated via numerical quadrature. To select a random set of quadrature points $\{h_n\}_{n=1}^N \in [h_n, h_f]$ a stratified sampling strategy is applied by uniformly drawing samples from evenly-spaced ray bins. 
A  pixel color   is approximated as:
\begin{align}
    C'(p,t) = \sum_{n=1}^{N}\mT'(h_n,t)&\alpha(h_n, t, \delta_n)\bc(\bp(h_n,t), \bd), \label{eq:sum}\\
    \quad \text{where} 
    \quad \alpha(h, t, \delta) =  1&-\exp(-\sigma(\bp(h,t))\delta), \\
    \quad \text{and} 
    \quad \mT'(h_n, t) = \exp&\left(-\sum_{{m}=1}^{n-1} \sigma(\bp(h_m,t)) \delta_m \right),
\end{align}
and $\delta_n = h_{n+1} - h_n$ is the distance between two quadrature points.

\subsection{Learning the Model}
\label{sec:learn}
The parameters of the canonical $\Psi_x$ and deformation $\Psi_t$ networks are simultaneously learned by minimizing the mean squared error with respect to the   $T$ RGB images $\{\bI_t\}_{t=1}^T$ of the scene  and their corresponding camera pose  matrices $\{\bT_t\}_{t=1}^T$. Recall that every time instant is only acquired by a single camera.

At each training batch, we first sample a random set of pixels $\{p_{t,i}\}_{i=1}^{N_s}$ corresponding to the rays cast from some camera position $\bT_t$ to some pixels $i$ of the corresponding RGB image $t$. We then estimate the colors of the chosen  pixels using Eq.~\eqref{eq:sum}. The training loss we use is   the mean squared error between the rendered and real pixels:
\begin{equation}
    \mL = \frac{1}{N_s}\sum_{i=1}^{N_s} \left \|\hat{C}(p,t) - C'(p,t) \right \|_2^2
    \label{eq:loss}
\end{equation}
where $\hat{C}$ are the pixels' ground-truth color.

\section{Implementation Details}
Both the canonical network $\Psi_x$ and the deformation network $\Psi_t$ consists on simple 8-layers  MLPs with ReLU activations. For the canonical network a final sigmoid non-linearity is applied to $\bc$ and $\sigma$. No non-linearlity is applied to $\bDelta\bx$ in the deformation network.

For all experiments we set the canonical configuration as the scene state at $t=0$ by enforcing it in Eq.~\eqref{eq:deformation}. To improve the networks convergence, we sort the input images according to their time stamps (from lower to higher) and then we apply a curriculum learning strategy where we incrementally add images with higher time stamps.

The model is trained with $400 \times 400$ images during $800k$ iterations with a batch size of $N_s=4096$ rays, each sampled $64$ times along the ray. As for the optimizer, we use Adam~\cite{kingma2014adam} with learning rate of $5e-4$, $\beta_1=0.9$, $\beta_2=0.999$ and exponential decay to $5e-5$. The model is trained with a single Nvidia\textsuperscript{\textregistered} GTX 1080 for 2 days.

\begin{figure*}
    \centering
    \includegraphics[width=\linewidth]{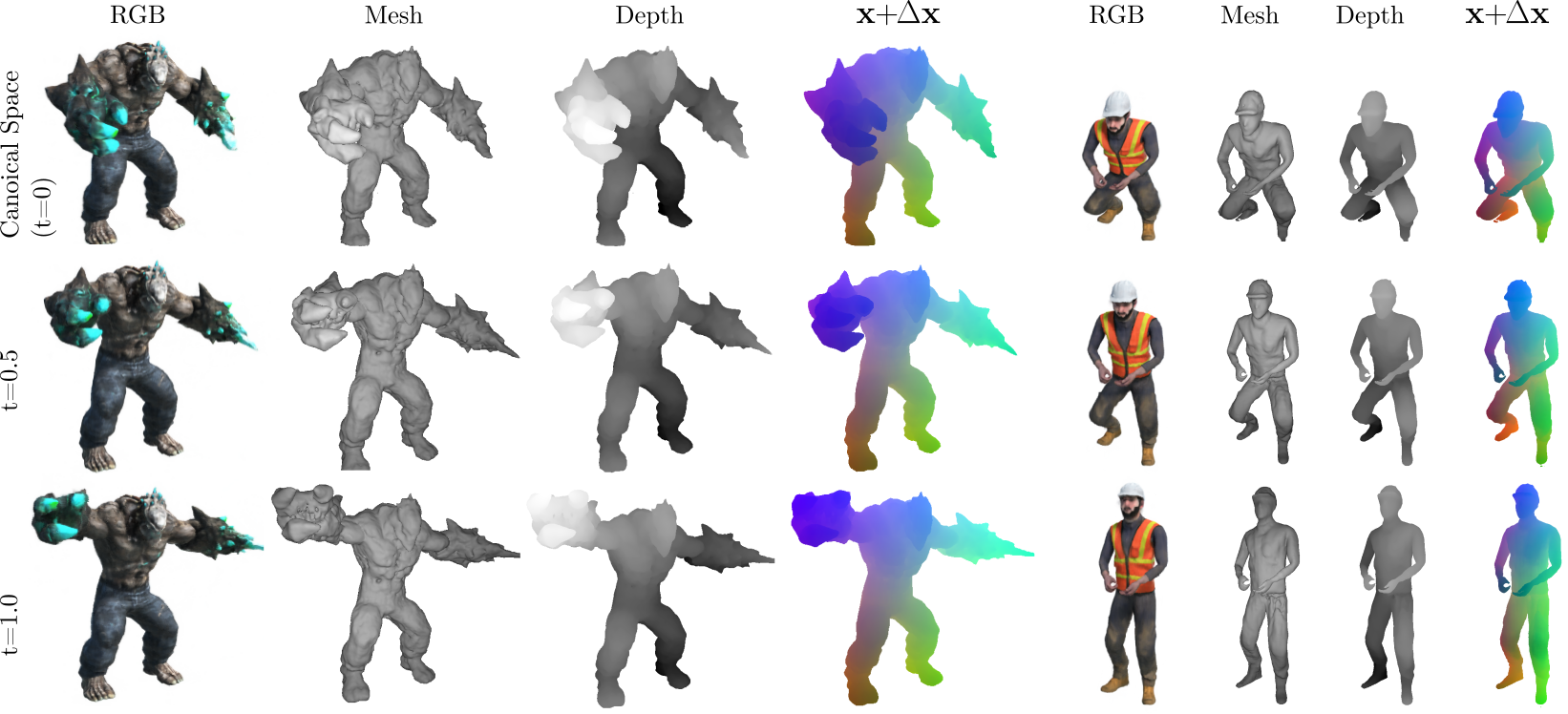}
    \vspace*{-2em} 
    \caption{\small{\textbf{Visualization of the Learned Scene Representation.} 
    Given a dynamic scene at a specific time instant, D-NeRF learns a displacement field $\Delta\bx$ that maps all points  $\bx$ of the scene to a common canonical configuration. The volume density and view-dependent emitted radiance for this configuration is learned and transferred to the original input points to render novel views. This figure represents, from left to right: the learned radiance from a specific viewpoint, the volume density represented as a 3D mesh and a depth map, and the color-coded points of the canonical configuration mapped to the deformed meshes based on $\Delta\bx$. The same colors on corresponding points indicate the correctness of such mapping.
    }}
    \vspace{-1.25em} 
    \label{fig:recon}
\end{figure*}
\begin{figure}
    \centering
    \includegraphics[width=\linewidth]{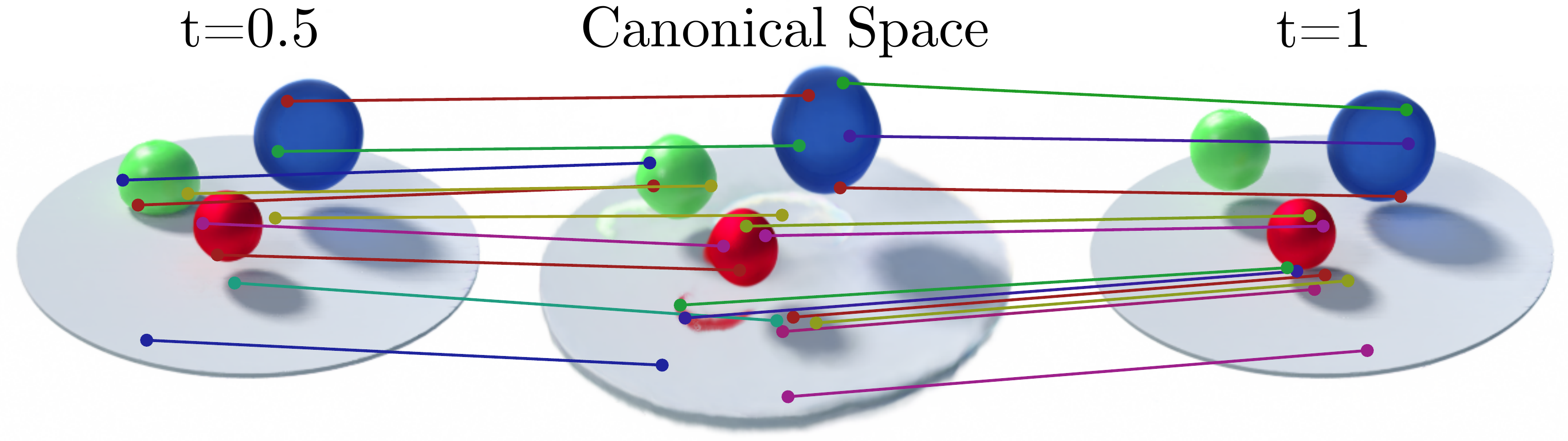}
    \vspace*{-2em} 
    \caption{\small{\textbf{Analyzing Shading Effects.} Pairs of corresponding points between the canonical space and the scene at times $t=0.5$ and $t=1$.}}
    \vspace{-1.75em} 
    \label{fig:shading}
\end{figure}

\section{Experiments}
This section provides a thorough evaluation of our system. We first test the main components of the model, namely the canonical and deformation networks (Sec.~\ref{sec:understand}). We then compare D-NeRF against NeRF and T-NeRF, a variant in which does not use the canonical mapping (Sec.~\ref{sec:compare}). Finally, we   demonstrate D-NeRF ability to synthesize novel views at an arbitrary time in several complex dynamic scenes (Sec.~\ref{sec:exp_time_angle}).

In order to perform an exhaustive evaluation we have extended NeRF~\cite{mildenhall2020nerf} rigid benchmark with eight  scenes containing dynamic objects under large deformations and realistic  non-Lambertian materials.  As in the rigid benchmark of~\cite{mildenhall2020nerf}, six are rendered from viewpoints sampled from the upper hemisphere, and two are rendered from viewpoints sampled on the full sphere. Each scene contains between 100 and 200 rendered views depending on the action time span, all at 800 × 800 pixels. We will release the path-traced images with defined train/validation/test splits for these eight  scenes.

\subsection{Dissecting the Model} 
\label{sec:understand}

This subsection provides insights about  D-NeRF behaviour when modeling a dynamic scene and analyze the two main modules, namely the canonical and deformation networks.

We initially evaluate the ability of the canonical network  to represent the scene in a canonical configuration.  The results of this analysis for two scenes are shown the first row of Fig.~\ref{fig:recon} (columns 1-3 in each case). The plots show, for the canonical configuration ($t=0$), the RGB image, the 3D occupancy network and the depth map, respectively. The rendered RGB image is the result of evaluating the canonical network on rays cast from an arbitrary camera position applying Eq.~\eqref{eq:sum}. To better visualize the learned volumetric density we transform it into a mesh applying marching cubes~\cite{lorensen1987marching}, with a 3D cube resolution of $256^3$ voxels. Note how   D-NeRF is able to model fine geometric and appearance details for  complex topologies and texture patterns, even when it was only trained with a set of sparse images, each under a different deformation.

In a second experiment we assess the capacity of the network to estimate consistent deformation fields that map  the canonical scene to the particular shape at  each input image. The second and third rows of Fig.~\ref{fig:recon} show the result of applying the corresponding translation vectors to the canonical space 
for $t=0.5$ and $t=1$. The fourth column in each of the two examples visualizes the displacement field, where the color-coded points in the canonical shape ($t=0$) at mapped to the different shape configurations at $t=0.5$ and $t=1$. Note that the colors are consistent along the time instants, indicating that the displacement field is correctly estimated.

Another question that we try to answer is how D-NeRF manages to model phenomena like shadows/shading effects, that is, how  
the model
can encode changes of appearance of the same point along time. We have carried an additional experiment to answer this.  In Fig.~\ref{fig:shading} we show a scene with three balls, made   of  very different  materials  (plastic --green--, translucent glass --blue-- and metal --red--). The figure plots pairs of corresponding points between the canonical configuration and the scene at a specific time instant.
D-NeRF is able to synthesize the shading effects by warping the canonical configuration. For instance, observe how the floor shadows are warped along time. Note that the points in the shadow of, \eg the red ball, at $t=0.5$ and $t=1$ map at different regions of the canonical space.

\begin{figure*}
    \centering
    \includegraphics[width=\linewidth]{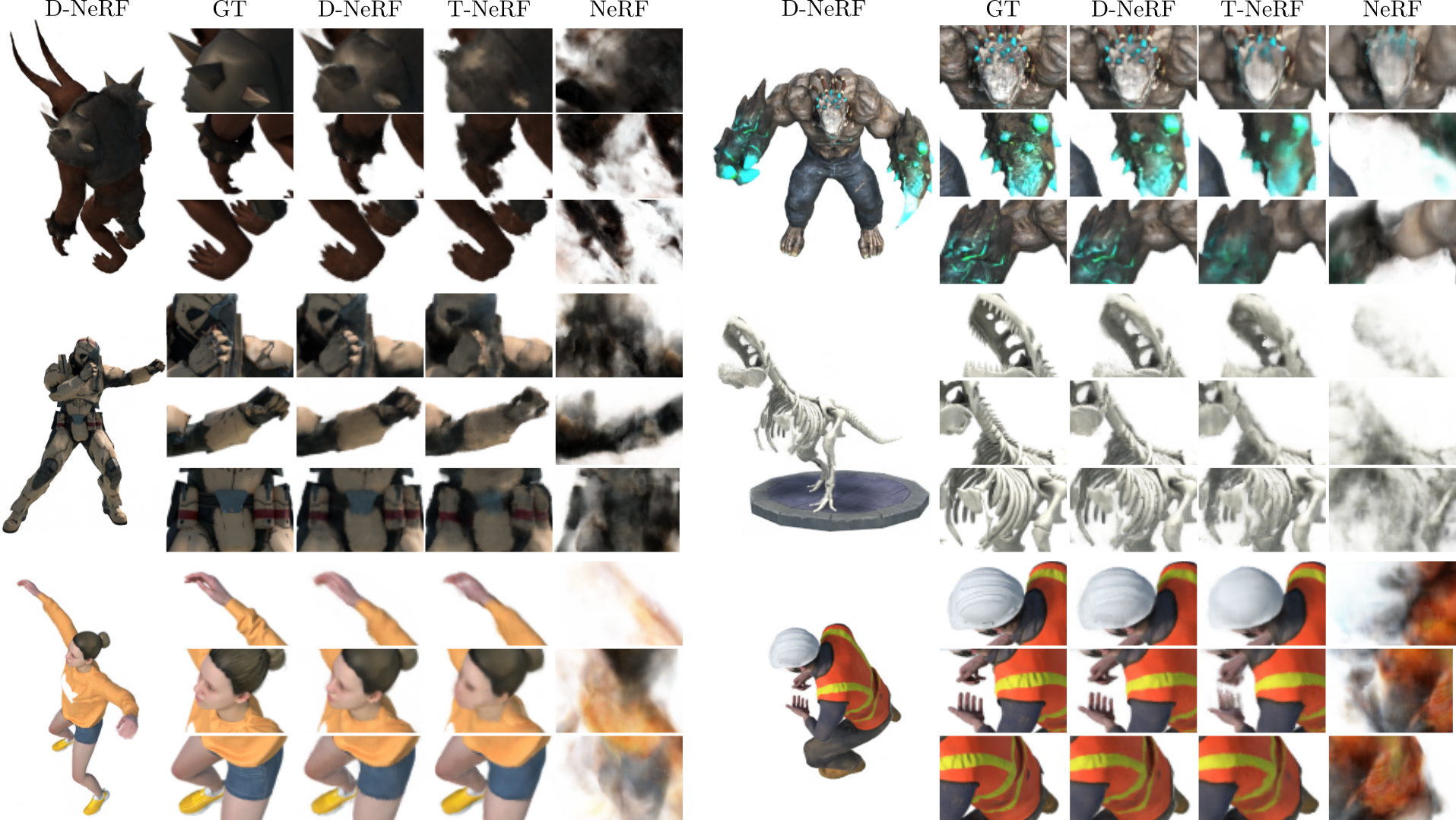}
    \vspace*{-2em} 
    \caption{\small{\textbf{Qualitative Comparison.} Novel view synthesis results of dynamic scenes. For every scene we show an image synthesised from a novel view at an arbitrary time by our method, and three close-ups for: ground-truth, NeRF, T-NeRF, and D-NeRF (ours).}}
    \vspace{-.5em} 
    \label{fig:comparison}
\end{figure*}
\begin{table*}[t!]
% \vspace{-.2em}
\setlength{\tabcolsep}{4pt} % general space between cols (6pt standard)
\centering
\resizebox{\textwidth}{!}{%
%\sisetup{detect-weight=true}
\begin{tabular}{lcccccccccccccccc}
\toprule
& \multicolumn{4}{c}{Hell Warrior} & \multicolumn{4}{c}{Mutant} & \multicolumn{4}{c}{Hook} & \multicolumn{4}{c}{Bouncing Balls} \\
Method 
& MSE$\downarrow$ & PSNR$\uparrow$  & SSIM$\uparrow$ & LPIPS$\downarrow$
& MSE$\downarrow$ & PSNR$\uparrow$  & SSIM$\uparrow$ & LPIPS$\downarrow$
& MSE$\downarrow$ & PSNR$\uparrow$  & SSIM$\uparrow$ & LPIPS$\downarrow$
& MSE$\downarrow$ & PSNR$\uparrow$  & SSIM$\uparrow$ & LPIPS$\downarrow$\\
\cmidrule(lr){1-1}  \cmidrule(lr){2-5} \cmidrule(lr){6-9} \cmidrule(lr){10-13} \cmidrule(lr){14-17}
    NeRF 
    & 44e-3 & 13.52 & 0.81 & 0.25
    & 9e-4 & 20.31 & 0.91 & 0.09
    & 21e-3 & 16.65 & 0.84 & 0.19
    & 1e-2 & 18.28 & 0.88 & 0.23 \\
    T-NeRF 
    & 47e-4 & 23.19 & 0.93 & 0.08
    & 8e-4 & 30.56 & 0.96 & 0.04
    & 18e-4 & 27.21 & 0.94 & \textbf{0.06}
    & 6e-4 & 32.01 & 0.97 & 0.04 \\
    D-NeRF 
    & \textbf{31e-4} & \textbf{25.02} & \textbf{0.95} & \textbf{0.06}
    & \textbf{7e-4} & \textbf{31.29} & \textbf{0.97} & \textbf{0.02} 
    & \textbf{11e-4}& \textbf{29.25} & \textbf{0.96} & 0.11
    & \textbf{5e-4} & \textbf{32.80} & \textbf{0.98} & \textbf{0.03} \\
\hline
% \\[.1em]
& \multicolumn{4}{c}{Lego} & \multicolumn{4}{c}{T-Rex} & \multicolumn{4}{c}{Stand Up} & \multicolumn{4}{c}{Jumping Jacks}  \\ Method
& MSE$\downarrow$ & PSNR$\uparrow$  & SSIM$\uparrow$ & LPIPS$\downarrow$
& MSE$\downarrow$ & PSNR$\uparrow$  & SSIM$\uparrow$ & LPIPS$\downarrow$
& MSE$\downarrow$ & PSNR$\uparrow$  & SSIM$\uparrow$ & LPIPS$\downarrow$
& MSE$\downarrow$ & PSNR$\uparrow$  & SSIM$\uparrow$ & LPIPS$\downarrow$\\  
\cmidrule(lr){1-1}  \cmidrule(lr){2-5} \cmidrule(lr){6-9} \cmidrule(lr){10-13} \cmidrule(lr){14-17}
    NeRF 
    & 9e-4 & 20.30 & 0.79 & 0.23
    & 3e-3 & 24.49 & 0.93 & 0.13
    & 1e-2 & 18.19 & 0.89 & 0.14
    & 1e-2 & 18.28 & 0.88 & 0.23 \\
    T-NeRF 
    & \textbf{3e-4} & \textbf{23.82} & \textbf{0.90} & \textbf{0.15} 
    & 9e-4 & 30.19 & 0.96 & 0.13
    & 7e-4 & 31.24 & 0.97 & 0.02
    & 6e-4 & 32.01 & 0.97 & 0.03 \\
    D-NeRF 
    & 6e-4 & 21.64 & 0.83 & 0.16 
    & \textbf{6e-4} & \textbf{31.75} & \textbf{0.97} & \textbf{0.03}
    & \textbf{5e-4} & \textbf{32.79} & \textbf{0.98} & \textbf{0.02}
    & \textbf{5e-4} & \textbf{32.80} & \textbf{0.98} & \textbf{0.03} \\

\bottomrule
\end{tabular}
}
\vspace*{-.75em} 
\caption{\small{\textbf{Quantitative Comparison.} We report MSE/LPIPS (lower is better) and PSNR/SSIM (higher is better).}}
\label{table:quant}
\vspace{-1em} 
\end{table*}

\subsection{Quantitative Comparison}
\label{sec:compare}
We next evaluate the quality of D-NeRF on the novel view synthesis problem and compare it against the original NeRF~\cite{mildenhall2020nerf}, which represents the scene using a 5D input  $(x,y,z,\theta,\phi)$,  and T-NeRF,  a straight-forward extension of NeRF in which the scene is represented by a 6D input $(x,y,z,\theta,\phi,t)$, without considering the intermediate canonical configuration of D-NeRF.

Table~\ref{table:quant} summarizes the quantitative results on the 8 dynamic scenes of our dataset. We use several metrics for the evaluation:   Mean Squared Error (MSE),   Peak Signal-to-Noise Ratio (PSNR),   Structural Similarity (SSIM)~\cite{wang2004image} and   Learned Perceptual Image Patch Similarity (LPIPS)~\cite{zhang2018perceptual}. 
In Fig.~\ref{fig:comparison} we show samples of the estimated images under a novel view for visual inspection. As expected, NeRF is not able to model the dynamics scenes as it was designed for rigid cases, and always converges to a blurry mean representation of all deformations. On the other hand, the T-NeRF baseline is able to capture reasonably well the dynamics, although is not able to retrieve high frequency details. For example, in Fig.~\ref{fig:comparison} top-left image it fails to encode the shoulder pad spikes, and in the top-right scene it is not able to model  the stones and cracks. D-NeRF, instead, retains high details of the original image in the novel views. This is quite remarkable, considering that each deformation state has only been seen from a single viewpoint.

\begin{figure*}
    \centering
    \includegraphics[width=\linewidth]{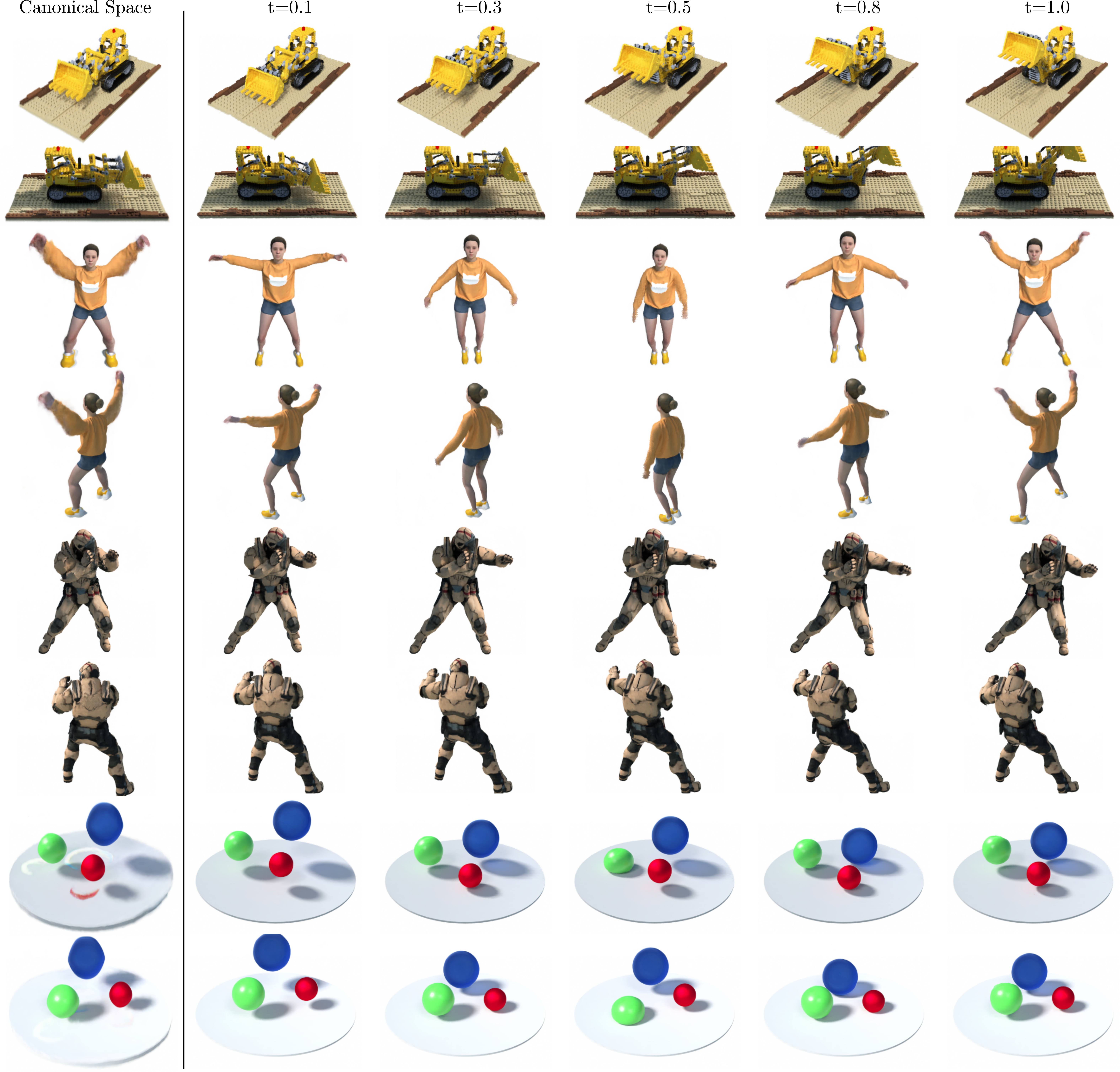}
    \vspace*{-2em} 
    \caption{\small{\textbf{Time \& View Conditioning.} Results of synthesising diverse scenes from two novel points of view across time and the learned canonical space. For every scene we also display the learned scene canonical space in the first column.}}
    \vspace*{-1.75em} 
    \label{fig:time_angle}
\end{figure*}

\subsection{Additional Results}
\label{sec:exp_time_angle}
We finally show additional results to showcase the wide range of scenarios that can be handled with D-NeRF.    Fig.~\ref{fig:time_angle}  depicts, for four  scenes, the images rendered at different time instants from two novel viewpoints. The first column displays  the canonical configuration. Note that we are able to handle several types of  dynamics: articulated motion in the {\em Tractor} scene;  human motion in the  {\em Jumping Jacks} and {\em Warrior} scenes; and asynchronous motion of several {\em Bouncing Balls}. 
Also note that the canonical configuration is a sharp and neat scene, in all cases, expect for the Jumping Jacks, where the two arms appear to be blurry. This, however, does not harm the quality of the rendered images, indicating that the network is able warp the canonical configuration so as to maximize the rendering quality. This is indeed consistent with Sec.~\ref{sec:understand} insights on how the network is able to encode shading.

\vspace{-0.5em}
\section{Conclusion}
\vspace{-0.4em}
We have presented D-NeRF, a novel neural radiance field approach for modeling dynamic scenes. Our method can be trained end-to-end from only a sparse set of images acquired with a moving camera, and does not require pre-computed 3D priors nor observing the same scene configuration from different viewpoints. The main idea behind D-NeRF is to represent   time-varying deformations with two modules: one that learns a canonical configuration, and another that learns the displacement field of the scene at each time instant  w.r.t. the canonical space.  A thorough evaluation demonstrates that  D-NeRF is able to  synthesise high quality novel views of scenes undergoing different types of deformation, from articulated objects to human bodies performing complex body postures. 

\footnotesize{
\paragraph{Acknowledgments}
This work is supported in part by a Google Daydream Research award and by the Spanish government with the project HuMoUR TIN2017-90086-R, the ERA-Net Chistera project IPALM PCI2019-103386 and María de Maeztu Seal of Excellence MDM-2016-0656.
Gerard Pons-Moll is funded by the Deutsche Forschungsgemeinschaft (DFG, German Research Foundation) - 409792180 (Emmy Noether Programme, project: Real Virtual Humans) }

%\noindent{\bf Acknowledgements}: 

{\small
\bibliographystyle{ieee_fullname}
\bibliography{references.bib}
}

\end{document}